# SOTIF Entropy: Online SOTIF Risk Quantification and Mitigation for Autonomous Driving

Liang Peng, Boqi Li, Wenhao Yu, Kai Yang, Wenbo Shao, and Hong Wang⊠, *Member, IEEE*

*Abstract*—Autonomous driving confronts great challenges in complex traffic scenarios, where the risk of Safety of the Intended Functionality (SOTIF) can be triggered by the dynamic operational environment and system insufficiencies. The SOTIF risk is reflected not only intuitively in the collision risk with objects outside the autonomous vehicles (AVs), but also inherently in the performance limitation risk of the implemented algorithms themselves. How to minimize the SOTIF risk for autonomous driving is currently a critical, difficult, and unresolved issue. Therefore, this paper proposes the "Self-Surveillance and Self-Adaption System" as a systematic approach to online minimize the SOTIF risk, which aims to provide a systematic solution for monitoring, quantification, and mitigation of inherent and external risks. The core of this system is the risk monitoring of the implemented artificial intelligence algorithms within the AV. As a demonstration of the Self-Surveillance and Self-Adaption System, the risk monitoring of the perception algorithm, i.e., YOLOv5 is highlighted. Moreover, the inherent perception algorithm risk and external collision risk are jointly quantified via SOTIF entropy, which is then propagated downstream to the decision-making module and mitigated. Finally, several challenging scenarios are demonstrated, and the Hardware-in-the-Loop experiments are conducted to verify the efficiency and effectiveness of the system. The results demonstrate that the Self-Surveillance and Self-Adaption System enables dependable online monitoring, quantification, and mitigation of SOTIF risk in real-time critical traffic environments.

## I. INTRODUCTION

ARTIFICIAL intelligence (AI) algorithms are widely adopted in autonomous driving systems to improve performance. However, AI algorithms typically provide black box solutions, and the randomness and unpredictability of those solutions will create inherent risk for autonomous vehicles (AVs). In addition, the risk of autonomous driving will also be triggered by complex external operational scenarios, such as extreme weather conditions and the stochastic behavior of road users. These types of risks fall within the scope of Safety of the Intended Functionality (SOTIF), which refers to the absence of unreasonable risk due to a hazard caused by a performance limitation, functional insufficiency, or reasonably foreseeable misuse [1].

SOTIF is relatively a new concept, but not a new issue; it is derived from Functional Safety (FuSa) [2]. Despite the application of AI algorithms in the autonomous driving system, FuSa was unable to address multiple critical issues, such as algorithm limitations and functional insufficiency. According to the reports of 3695 takeovers collected by the California Department of Motor Vehicles for the 2020 road test of BMW, Toyota, etc., 90.31% of the disengagements were due to software system SOTIF issues [3]. Among them, problems associated with perception, prediction, and planning accounted for 13.94%, 3.30%, and 35.23%, respectively of the SOTIF issues.

SOTIF issues have had severe repercussions. One typical SOTIF case is that the perceptual system of one intelligent vehicle failed to distinguish a white truck from the sky on a sunny day in the year 2016, resulted in a fatal collision [4]. Similarly, this kind of intelligent vehicles hit three more white trucks in Florida, Taiwan, and Detroit due to the perceptual system's performance limitations. One intelligent vehicle failed to prevent a cyclist from crossing the street illegally at night, which is another example of a fatal SOTIF case in the year 2018 [5]. This case was triggered by the functional insufficiency of its decision-making system, which failed to account for jaywalking pedestrians and respond to fluctuating perception results.

To sum up, the SOTIF issue necessitates two conditions: trigger conditions and system performance limitations or functional insufficiencies. Taking the perceptual subsystem as an instance, the first factor, which includes weather conditions and road conditions, is highly relevant to the operational design domain (ODD). Weather conditions, such as rainy, foggy, and snowy weather, as well as direct sunlight, will significantly degrade the performance of the perception system, leading to

This paragraph of the first footnote will contain the date on which you submitted your paper for review, which is populated by IEEE. The authors would like to appreciate the contributions of the perception task group of the CAICV-SOTIF technical alliance in China. This work was supported in part by the National Natural Science Foundation of China Project U1964203 and Project 52072215; and in part by the National Key Research and Development Program of China under Grant 2020YFB1600303. *(Corresponding author: Hong Wang).*

Liang Peng, Wenhao Yu, Wenbo Shao and Hong Wang are with Sate Key Laboratory of Automotive Safety and Energy, School of Vehicle and Mobility in Tsinghua University, Beijing 100084, China (e-mail: peng-l20@mails.tsinghua.edu.cn; wenhaoyu@mail.tsinghua.edu.cn; swb19@mails.tsinghua.edu.cn; hong_wang@mail.tsinghua.edu.cn).

Boqi Li is with the Department of Mechanical Engineering, University of Michigan, Ann Arbor, MI 48109, USA (e-mail: boqili@umich.edu).

Kai Yang is with the College of Automotive Engineering, Chongqing University, Chongqing, China (e-mail: kaiyang0401@gmail.com).

Digital Object Identifier.

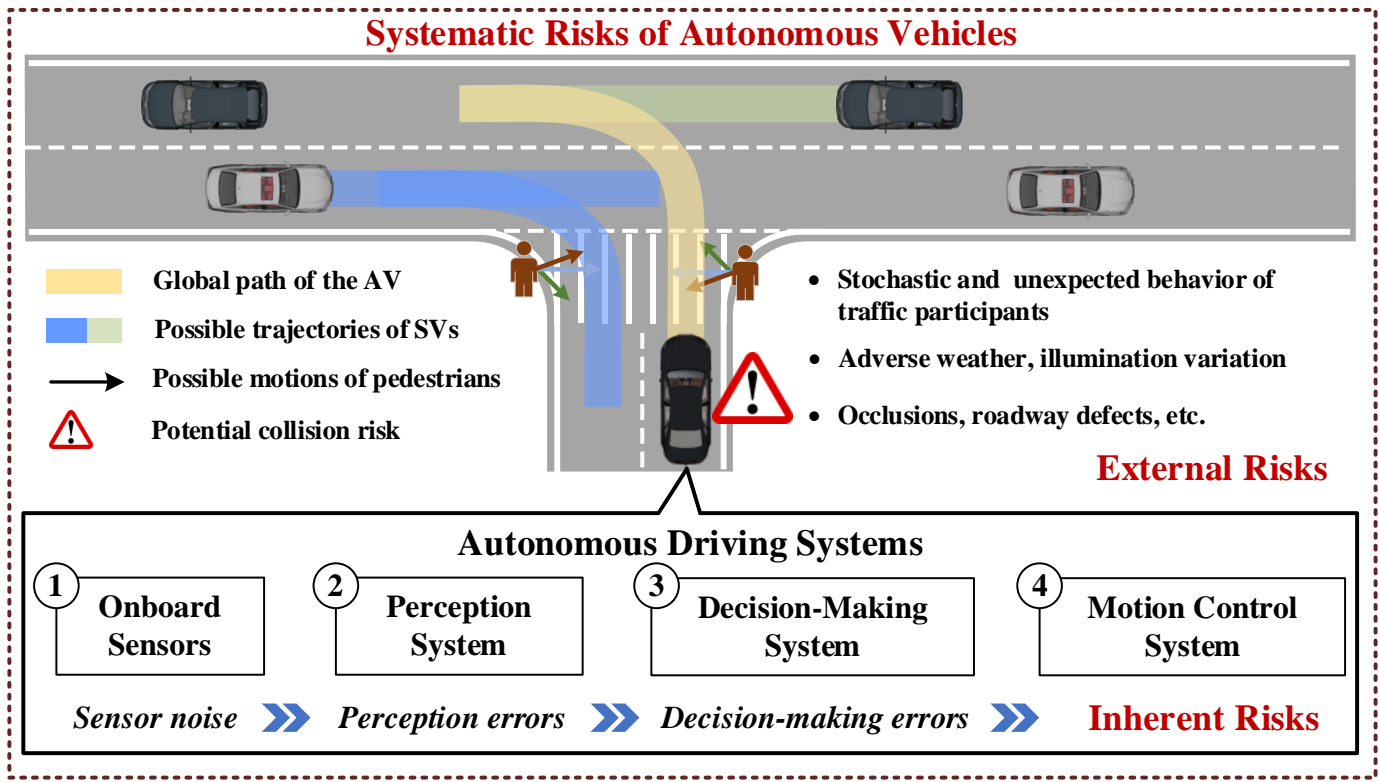

**Fig. 1** Composition of the overall SOTIF risks of the autonomous vehicles, including external risks and inherent risks.

SOTIF issues. In addition, road conditions, such as ropes, drawbars, falling objects from surrounding vehicles (SVs), and other unanticipated occurrences, will pose significant obstacles for the autonomous driving system. The second necessary factor for SOTIF issues is related to the amount of data used to train the algorithms. Typically, the system's performance degrades when exposed to long-tail and out-of-distribution scenarios.

In the dynamic operational environment of autonomous driving, many sources can trigger SOTIF threats. Especially, several of them can directly cause the perception system to misinterpret traffic scenarios, affecting the subsequent modules. Based on the theoretical examination of the physical and algorithmic principles of perception systems, Wu *et al*. formed a comprehensive list of factors and generated edge cases at the semantic level [6]. Further, Xing *et al*. proposed an analysis framework of perception system trigger conditions based on the chain of events model and another framework based on trigger conditions to support the safety analysis and verification of AVs [7].

Regarding the whole system of AVs, extreme weather, poor lighting, road defects, temporary obstacles, unexpected target object behavior, and disturbing objects like animals and falling objects are all considered as trigger sources. As shown in Fig. 1, SOTIF risks triggered by these trigger sources can be divided into external and inherent risks. The external risk is the explicit possibility of colliding with entities on the road, such as target objects, obstacles, and road edges. The risk modeling of an external entity is often carried out based on its category, position, and velocity relative to the ego vehicle. The inherent risk is the implicit risk of algorithm performance degradation within the autonomous driving system, which is associated with the dynamic operational environment and the limitations of sensors and algorithms.

Most existing autonomous driving systems only take external risk into account, ignoring the inherent risk modeling of AI algorithms. In practice, the entity category, relative position, relative velocity, and other information required for external risk modeling depend on upstream AI algorithms, whose unpredictability and black-box characteristics may result in the failure of external risk modeling. For instance, if the perceptual system fails to detect a target object, the vehicle will not respond. Therefore, the overall SOTIF risks, which includes both inherent and external risks, must be systematically considered. However, fundamental theories and systematic solutions are currently lacking for preventing the SOTIF risks in real-time. The detailed contributions of this paper are hence summarized as follows:

1) The Self-Surveillance and Self-Adaption System is a proposed framework for monitoring, quantification, and mitigation of both inherent and external risks of AVs.

2) This study investigates a demonstration of the proposed Self-Surveillance and Self-Adaption System for processing perception algorithm risk. The risk monitoring of the perception algorithm is highlighted in this paper. It includes monitoring the risk of YOLOv5 and quantifying the associated risk with SOTIF entropy, which can be mitigated with uncertainty-aware decision-making strategies.

3) A Perceptual Uncertainty-Aware Decision-Making (PUADM) method is proposed to mitigate the risks propagated

by the perception algorithm. The effectiveness and efficiency are verified via Hardware-in-the-Loop experiments under multiple critical scenarios.

The remainder of this paper is organized as follows. The related work of quantification and mitigation of SOTIF risks for autonomous driving systems employing AI algorithms is presented in Section 2. Section 3 outlines the Self-Surveillance and Self-Adaption System. Section 4 describes a demonstration of handling the inherent perceptual risk in detail, including entropy quantification and risk mitigation modules. In Section 5, comprehensive experiments are compared and analyzed under typical perceptual SOTIF-related scenarios, and in Section 6, conclusions are drawn.

## II. RELATED WORK

According to the preceding analysis, it is essential to minimize the SOTIF risk systematically. In this regard, the systematic solutions for reducing the risk of FuSa are first examined. Then, the research on modeling the inherent SOTIF risk is investigated from the perspective of handling the uncertainty of AI algorithms.

### A. System Solution of Safety Guidelines for AVs

Krzysztof Czarnecki *et al*. proposed an approach and architectural design for modifying the runtime representation of ODD based on system capability in [8]. This architecture consisted of a system layer and a supervisory layer. The system layer is responsible for executing the dynamic driving task, while the supervisory layer is responsible for monitoring the system and interacting with the system layer to respond to impairments. The proposed architecture could deal with most issues associated with running out of ODD in the functional safety concept, but not the SOTIF issues specifically.

In 2012, the project Stadpilot, which aims to achieve fully autonomous driving on the innercity ring road of Braunschweig, first proposed a systematic solution. Andreas Reschka *et al*. proposed a surveillance and safety system based on performance criteria and functional degradation for AVs in [9]. The surveillance system collected data from hardware sensors and software systems. Degradation actions and safety maneuvers would be executed according to the collected data, such as rain amount, temperature, sideslip angle, tire velocities, etc., to keep the vehicle safe.

However, the introduced systematic solutions were incapable of addressing SOTIF issues associated with trigger conditions, such as extreme weather, nor those associated with algorithm performance limitations and functional inefficiencies. Future systematic solutions for AVs must therefore consider monitoring the limitations of algorithms and making self-adaptive safe decisions.

### B. Uncertainty Monitoring and Risk Quantification

Recent advancements have made the autonomous driving system, particularly the perception subsystem, highly dependent on artificial intelligence techniques. However, it is notoriously challenging to guarantee their safety, and they may exhibit unexpected behavior, which is unacceptable for safety-critical applications. To mitigate this issue, the systematic solution described in [8] suggested designing a supervisory layer to monitor the AI algorithms and interact in real-time with the system layer. Once a degraded performance is detected, the system layer will receive an early warning and modify its behavior accordingly. For example, for monitoring perceptual object detection algorithms, the method of predicting per-frame mean Average Precision (mAP) proposed by Rahman *et al*. can monitor performance degradation without the need for ground truth data. However, its real-time performance and false alarm rate have yet been verified [10].

As previously emphasized, the black-box nature of AI algorithms is one of the most important factors contributing to SOTIF risk. In recent years, this characteristic has been studied primarily via uncertainty estimation. The uncertainty of the algorithm can indicate the output's effectiveness and is suitable for monitoring. Perception, in particular, is a safety-critical function of AVs, and AI algorithms play a crucial role in its implementation. Perceptual uncertainties usually result in SOTIF issues in the nominal performance of learning-based algorithms. Moreover, failures in perception usually result in unsafe maneuvers because the planning and control modules rely heavily on perception results.

TABLE I
TYPICAL METHODS OF UNCERTAINTY ESTIMATION

| Method | Basis | Uncertainty Type | Pros and Cons |
|---|---|---|---|
| VI | Bayesian theory | Epistemic uncertainty | Functional analysis [16]; precise theoretical derivation; high computational complexity. |
| MCMC | Bayesian theory | Epistemic uncertainty | Markov process [17]; convenient approximate inference; high computational complexity. |
| LA | Bayesian theory | Epistemic uncertainty | Gaussian posterior [15]; simple distribution form; high computational complexity. |
| MCD | Sampling-based | Epistemic uncertainty | Bernoulli weights [18]; easy to implement by dropout; need more samples than DE. |
| DE | Sampling-based | Epistemic uncertainty | Bootstrap combination [25]; efficient parallel inferences; only for white-box networks. |
| GOM | Gaussian distribution | Aleatoric uncertainty | Gaussian output [29]; efficient for single inference; need modifying layers and losses. |
| DOM | Dirichlet distribution | Aleatoric uncertainty | Dirichlet output [31]; feasible for black-box networks; need sampling like MCD. |
| DER | Evidence theory | Both | Evidential Prior [35]; efficient and scalable learning; complex evidence theory. |
| EP | Propagation mechanism | Both | Variance propagation [36]; limited modifications to layers; complex propagation mechanism. |

Kendall *et al*. analyzed the role of uncertainty estimation in image classification and semantic segmentation, separating the output prediction uncertainty of deep neural networks into epistemic and aleatoric uncertainties [11]. The epistemic uncertainty refers to the algorithm's inherent cognitive ability, while the aleatoric uncertainty reflects the influence of input noise on the algorithm. Krzysztof *et al*. provided a summary of the perceptual uncertainties that may lead to perception failures during the development and operation phases of learning-based models, such as labeling uncertainty, model uncertainty, operational domain uncertainty, etc. [12]. The modeling and protection of inherent risk can be achieved by estimating and monitoring the AI algorithmic uncertainties.

The Bayesian model provides a mathematical framework for estimating uncertainties, but the increase in computation costs is a drawback [13]. Furthermore, due to the non-linearity of deep neural networks, the accurate posterior inference is difficult to achieve [14]. In terms of theoretical research, Laplace Approximation (LA), Variational Inference (VI), Markov Chain Monte Carlo (MCMC), and other similar methods can simplify the inference process of the Bayesian model. Nevertheless, the computational efficiency of these methods is still insufficient [15], [16], [17]. As research has progressed, more practical approaches have been proposed. Tabel I summarizes the pros and cons of various uncertainty estimation methods.

For epistemic uncertainty, Gal *et al*. demonstrated that utilizing the Monte Carlo Dropout (MCD) method in deep neural networks could be interpreted as a Bayesian approximation of Gaussian processes [18], [19], [20]. Peng *et al*. utilized the MCD method to obtain probabilistic object detectors and estimate the uncertainty for each object's bounding box (regression value) and category (classification value) [21], [22]. However, each model's weights should be sampled at the outset of the inference procedure, which would increase computational costs [23]. Deep Ensembles (DE) is another Bayesian approximation method that trains an ensemble of deterministic models with randomly initialized weights and random shuffling of data during training, as demonstrated by Osband *et al*. [24], [25], [26]. Several studies trained ensembles of object detection models with different architectures to complement each other, reduce the number of missing objects, and enhance the mAP [27], [28]. However, the DE method for estimating the uncertainty of an object detector for safety evaluation has not been extensively studied. This paper adopts the DE technique because it is simple to be implemented, conducive to be parallel operated, and has the potential to satisfy real-time requirements. This paper adopts it in the uncertainty estimation and entropy quantification of the perception algorithm.

For aleatoric uncertainty, different distributions are usually imposed on the network output as the prior. The Gaussian Output Modeling (GOM) method typically assumes that each class score of the network output follows an independent Gaussian distribution, which does not require sampling and is widely adopted [29], [30]. In addition, Mena *et al*. assumed that the network output follows the Dirichlet distribution and created wrappers to estimate the uncertainty of black-box algorithms based on this assumption [31], [32]. However, the real-time performance of this Dirichlet Output Modeling (DOM) method cannot be guaranteed because it requires sampling and introduces an additional white-box network [33].

In general, different methods can be combined to quantify the overall uncertainty of a prediction. For example, Harakeh *et al*. used the MCD method and the GOM method, respectively, to measure epistemic uncertainty and aleatoric uncertainty [34]. In addition, some studies estimate both epistemic and aleatoric uncertainty at once. Amini *et al*. proposed the Deep Evidential Regression (DER) method to directly measure the prediction uncertainty, which places evidential priors over the original Gaussian likelihood function [35]. In addition, the Error Propagation (EP) method is also attracting practical applications. The batch normalization layers are viewed as noise-injection procedures, and the activation layers are transformed into layers of uncertainty propagation [36], [37]. This sampling-free method can propagate the uncertainty estimated in injection layers to the output layer via propagation layers in a single inference, which has high computational efficiency.

Given that the autonomous driving system comprises multiple modules, it is essential to quantify the risks based on the monitored uncertainties and propagate them downstream. The position and velocity of an object are continuous and convenient for the decision-making module to consider. Kahn *et al*. propagated the prediction results with uncertainty to the planning module, which can minimize dangerous collisions [38], [39]. However, the classification of an object is categorical, and the modeling of driving safety fields with traversable and untraversable obstacles is very different [40]. For instance, suppose a pedestrian is misclassified as a traffic cone and the perception module outputs only the category to the decision-making module, the vehicle may take dangerous actions with confidence [41], [42]. Ivanovic *et al*. took the class uncertainty into account. The prediction module receives all class scores instead of a single category to predict trajectories for various possible categories and produce a final prediction [43], [44]. Nonetheless, the uncertainty of the perception algorithm has not been evaluated in this study.

III. SELF-SURVEILLANCE AND SELF-ADAPTION SYSTEM

As depicted in Fig. 2, this section proposes a systematic solution for monitoring, quantification, and mitigation of both inherent and external risks of AVs. As avoiding obstacles is the basic task of AVs, the external risk $R_e$ can be modeled by the well-studied localization, perception, prediction, and planning algorithms. Moreover, the inherent risk $R_i$ can be modeled by monitoring the operational status of AI algorithms through the Self-Surveillance and Self-Adaptive Safety System.

As previously emphasized, the safety of AI algorithms is particularly important for SOTIF. Therefore, each AI-powered module in the autonomous driving system must be monitored.

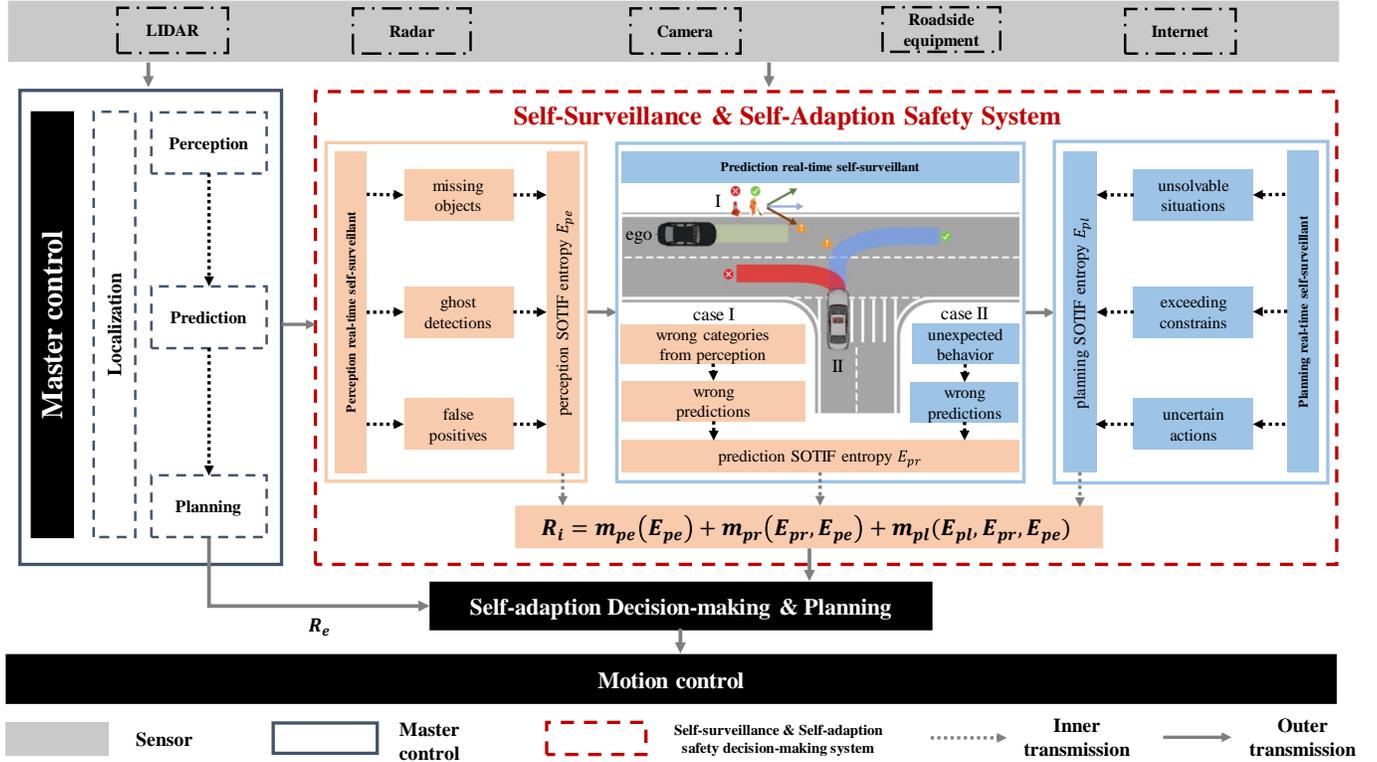

**Fig. 2** The framework of the Self-Surveillance and Self-Adaption Safety System.

Specifically, the inherent risk $R_i$ mentioned in this paper includes three aspects, *i.e.*, the inherent risks of the perception module, prediction module, and planning module, denoted as $E_{pe}$, $E_{pr}$, and $E_{pl}$. The inherent risk of the perception subsystem $E_{pe}$ refers to the risk of missing objects, ghost defections, and false positives of perception algorithms. The inherent risk of the prediction subsystem $E_{pr}$ refers to the wrong predictions of prediction algorithms confronted with unexpected behavior or wrong perception results. The inherent risk of the planning subsystem $E_{pl}$ refers to the unsolvable situations of rule-based decision-making algorithms, solutions exceeding constraints, and the uncertain actions of learning-based planning algorithms. Note that the risk of the module upstream will also affect the modules downstream [45], [46]. For instance, when the perception algorithm transmits incorrect object categories, the prediction algorithm will likely make incorrect predictions, thereby increasing the inherent risk. In summary, the implicit expression of the overall inherent risk is provided by the mappings related to the inherent risks of the subsystems.

Consequently, the overall risk of the AV system consists of inherent risk $R_i$ from AI algorithms and external collision risk $R_e$. If the monitored and quantified overall risk falls within an acceptable range, the self-adaptive decision-making and planning algorithm will take it into account to implement motion control to mitigate the risk. However, if the overall risk is too high, it is recommended to activate the degradation mechanism or simply pull over.

Given that AI algorithms are widely adopted in the perception module, which is located upstream of the autonomous driving system so plenty of trigger conditions intuitively act on it. To demonstrate the effectiveness of the proposed framework, a demonstration of perception algorithm risk processing is investigated. This paper focuses on the risk monitoring and protection measures of the perception algorithm, which is verified from the perspectives of perception performance and overall system performance.

IV. A Demonstration of Handling the Inherent Perceptual Risk

The perceptual SOTIF risk of AVs may be triggered by environmental factors such as rain, snow, fog, and poor lighting conditions, as well as object characteristics such as unusual appearances or postures. In the aforementioned circumstances, learning-based perception algorithms often generate incorrect detection results. The causes can be summed up from two perspectives: the deterioration of sensor capabilities and the limited cognitive capacity of networks. However, current autonomous driving systems assume that the perceptual results are 100% accurate and transmit the results directly to the modules downstream, which could result in dangerous decisions and even fatal accidents [4], [5]. Therefore, it is crucial for safe driving to monitor the operational states of learning-based perception algorithms, quantify the associated risk, and carefully mitigate it in modules downstream.

This section displays a demonstration of the Self-Surveillance and Self-Adaption Safety System. In this demonstration, the perception module uses the YOLOv5 object detection algorithm, and the planning module is based on Model Predictive Control (MPC). Specifically, this paper utilizes Deep Ensembles to estimate the epistemic uncertainty of the AI perception algorithm to realize the self-monitoring of the perception module. In addition, the concept of entropy is used to quantify the monitored perceptual risk. The quantified perception SOTIF entropy ultimately bypasses the prediction module. It is directly propagated to the planning module in real-time to generate safer trajectories where the overall risk is mitigated.

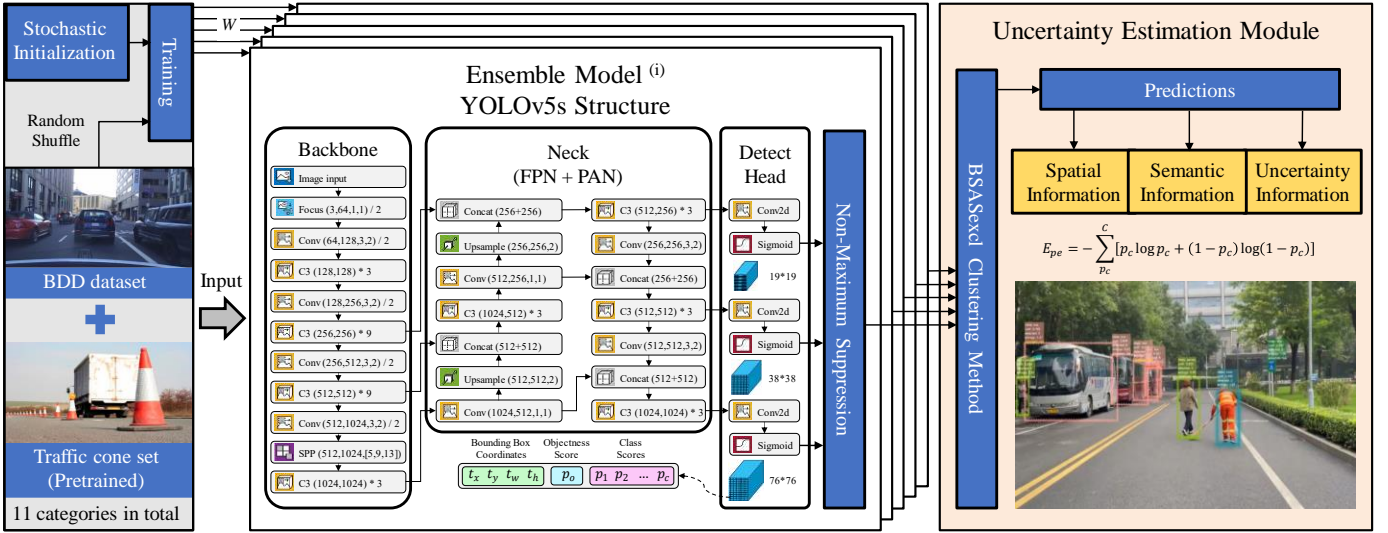

**Fig. 3** Modeling the epistemic uncertainty of the AI perception algorithm.

*A. Monitoring the Perceptual Epistemic Uncertainty*

The Deep Ensembles method is a sampling-based method for modeling epistemic uncertainty. The ensemble's object detection networks are deterministic and adhere to the same architecture with slightly different weights. Moreover, it has been demonstrated that an ensemble of five networks is sufficient for operation, allowing different GPUs or threads to be used conveniently for parallel inference [25]. Therefore, this method only introduces the time cost of post-processing in theory, which offers a potential practical implementation opportunity in the autonomous vehicle.

Applying the Deep Ensembles method, $T$ object detectors are trained with the same training process and dataset. However, different random number seeds are used to generate an ensemble $E = \{O_1, O_2, ..., O_T\}$, as shown in Fig. 3. For each input frame image $I$, a set of predictions $P = \{P_1, P_2, ..., P_T\}$ can be obtained through different object detectors, where each $P_i$ is the result after independently running the Non-Maximum Suppression (NMS) [30]. As shown in Algorithm 1, a set of clusters $L = \{L_1, L_2, ..., L_N\}$ can be obtained from $P$ through the Basic Sequential Algorithm Scheme with intra-sample exclusivity (BSAS excl.) based on the Intersection over Union (IoU) and the winning label (WL), in which each $L_i$ corresponds to a detected object and contains at most $T$ sampling tensors [28], [47]. The mean of all sample tensors for each cluster $L_i$ is treated as the final perception result. Then, the variances of coordinates of the bounding boxes are used to approximate the spatial uncertainty.

**Algorithm 1** Basic Sequential Algorithm Scheme with Intra-Sample Exclusivity

**Input:** $Affinity = \{IoU \& WL\}$, threshold $\theta_{aff}$, a set of predictions $P = \{P_1, P_2, ..., P_T\}$;

**Output:** A set of clusters $L = \{L_1, L_2, ..., L_N\}$;

1: Create a cluster for each object box in $P_1$;
2: **for** $i \in \{2, 3, ..., T\}$
3:   set $excl\_flag = \mathbf{0_n}$, n is the current number of clusters
4:   **for** each object box $B_j$ in $P_i$
5:     **for** $k \in \{1, 2, ..., n\}$
6:       **if** $Affinity(B_j, L_k) \geq \theta_{aff}$ **and** $excl\_flag(k) = 0$ **then**
7:         put $B_j$ into $L_k$, $excl\_flag(k) = 1$
8:       **if** $B_j$ has not been processed yet **then**
9:         $n = n+1$, create a new cluster $L_n$ for $B_j$
10: **return** $L$;

*B. Quantifying the Perception SOTIF Entropy*

In probabilistic object detection, the sample mean and variance are frequently employed to approximate the corresponding posteriori distribution statistics. Meanwhile, Shannon entropy is also a popular metric for estimating the quality of predictive uncertainty in classification tasks [48]. For a category label $y$ of $C$ categories, the Shannon entropy $H$ is measured by:

$$H = -\sum_{c=1}^{C} p(y=c|x,D) log(p(y=c|x,D)) \qquad (1)$$

where $x$ represents the input data and $D$ is the training dataset. $H$ reaches a minimum value $H_{min} = 0$ when the network is completely certain in its prediction, i.e., $p(y = c | x, D) = 0 \ or \ 1$. When the prediction of the network follows a uniform distribution, i.e., $p(y = c | x, D) = \frac{1}{C}$, $H$ reaches a maximum value, i.e., $H_{max} = log(C)$.

The above metric is applicable in single-label object detection, where the probability vector output is subject to category distribution, or $\sum_{c=1}^{C} p(y = c | x, D) = 1$. While YOLOv5 supports multi-label object detection, which has higher fault tolerance. It uses $C$ independent logistic classifiers to predict the probabilities that the result belongs to a certain category. In summary, this paper quantifies the perception SOTIF entropy $E_{pe}$ of the YOLOv5s network as:

$$p_c = p(y = c | x, D) \approx \frac{1}{T}\sum_{t=1}^{T} p(y = c | x, W_t) \quad (2)$$

$$E_{pe} = -\sum_{p_c}^{C}(p_c log p_c + (1-p_c)log(1-p_c)) \quad (3)$$

where $T$ represents the number of networks in the ensemble, $W_t$ denotes the corresponding weights of the $t$-th network, and $C$ is the number of categories. For the probability $p_c$ corresponding to the $c$-th category, $[p_c, 1-p_c]$ is regarded as the probability vector output from the binary classification problem, and its Shannon entropy $H_c$ takes $p_c = \frac{1}{2}$ as the axis of symmetry and is on a convex shape. Hence $E_{pe}$ reaches a minimum value $E_{pe\_min} = 0$ when the ensemble is completely certain in its all predictions, i.e., $p_c = 0 \ or \ 1$ for each $c$. When the predictions of the ensemble are all ambiguous, i.e., $p_c = \frac{1}{2}$ for each $c$, $E_{pe}$ reaches a maximum value, i.e., $E_{pe\_max} = Clog2$. Combined with (2), it can be concluded that $E_{pe}$ tends to be large when there are errors or disagreements among the networks, no matter whether they are caused by environmental factors or object attributes. Therefore, $E_{pe}$ can be used for self-monitoring and generating warning signals.

The output of the perception algorithm modified in this paper includes three types of information, namely, spatial information with uncertainty, semantic information with uncertainty, and comprehensive uncertainty information. The spatial information reflects the uncertainty of the bounding box positioning. As shown in Fig. 3, two dashed boxes are added around each bounding box whose corner coordinates are three times the standard deviation apart. The semantic information reflects the uncertainty of the final perception results. The category with the highest probability within the class probability vector is referred to as the winning label, and the probability associated with it is recorded as the confidence score. The comprehensive uncertainty information reflects the macroscopic uncertainty results of the cluster corresponding to the object, including the number of object detectors that have detected the object, the prediction entropy of the final perception result, and the uncertainty level as determined by the prediction entropy. When only a portion of the ensemble detects an object, it may be a ghost detection or a missing object for a particular network. Consequently, an additional penalty is imposed on this object's prediction entropy:

$$E_{pe}^* = E_{pe} \times (1 + f_p \times (T - d)) \quad (4)$$

where $E_{pe}^*$ refers to the final perception SOTIF entropy, $f_p$ represents the penalty factor, and $d$ refers to the number of networks that detects the object in the ensemble. Then, to output the monitoring results, three entropy levels are defined as:

$$l_u = \begin{cases} 0 & if \ E_{pe}^* < \theta_{lm} & low, \ normal \\ 1 & if \ \theta_{lm} \le E_{pe}^* < \theta_{mh} & medium, \ caution \\ 2 & if \ \theta_{mh} \le E_{pe}^* & high, \ warning \end{cases} \quad (5)$$

where $l_u$ represents the uncertainty level, $\theta_{lm}$ refers to the uncertainty threshold between low and medium uncertainty, and $\theta_{mh}$ denotes the uncertainty threshold between medium and high uncertainty. If an object has a high level of uncertainty, a warning will be generated to alert the driver or subsequent planning module to pay attention to it.

*C. Modeling and Mitigating the Overall Risk*

As previously emphasized, it is essential to account for perceptual uncertainty in the downstream module, i.e., the decision-making system. Consequently, a Perceptual Uncertainty-Aware Decision-Making (PUADM) method is proposed to handle the risks propagated by the perception system.

*a. Potential Field Incorporating Class Uncertainty*

First, it is known that the class of road objects directly impacts the decision-making process of AVs. For instance, AVs should be cautious around pedestrians walking along the side of the road and maintain a safe lateral distance from them, as pedestrians' behaviors are typically stochastic. However, this is not necessary for traffic cones placed on the side of the road. The following description of the potential field method will be used to address this issue.

$$PF_{U_{i,c}}(X, Y) = a_c e^{\beta_{i,c}} \quad (6)$$

with

$$\beta_{i,c} = -(\frac{((X-P_{x,i})cos\theta_i + (Y-P_{y,i})sin\theta_i)^2}{2\lambda_{x_{i,c}}^2} \\ + \frac{(-(X-P_{x,i})sin\theta_i + (Y-P_{y,i})cos\theta_i)^2}{2\lambda_{y_{i,c}}^2})^{b_c} \quad (7)$$

$$\lambda_{x_{i,c}} = L_{x_{i,c}} + E_{x_{i,c}} \quad (8)$$

$$\lambda_{y_{i,c}} = L_{y_{i,c}} + E_{y_{i,c}} \quad (9)$$

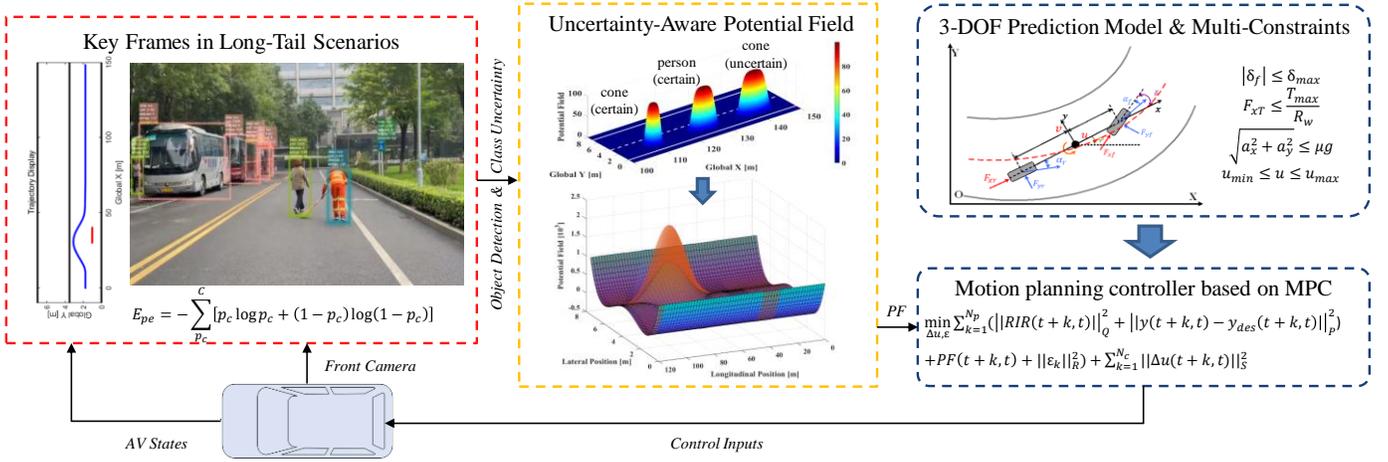

**Fig. 4** Perceptual uncertainty-aware safe decision-making algorithm.

$$E_{x_{i,c}} = \begin{cases} 0 & if\ l_u = 0 \\ L_{x_{i,person}} - L_{x_{i,c}} & if\ l_u = 1 \\ L_{x_{i,person}} - L_{x_{i,c}} + U_x & if\ l_u = 2 \end{cases} \quad (10)$$

$$E_{y_{i,c}} = \begin{cases} 0 & if\ l_u = 0 \\ L_{y_{i,person}} - L_{y_{i,c}} & if\ l_u = 1 \\ L_{y_{i,person}} - L_{y_{i,c}} + U_y & if\ l_u = 2 \end{cases} \quad (11)$$

where subscripts $i$ and $c$ represent objects' ID and category, respectively, $a_c$ and $b_c$ are predefined intensity and shape factors of potential field subject to the object category $c$, respectively, $X$ and $Y$ represent the coordinates of ego vehicle, $P_{x,i}$ and $P_{y,i}$ are the coordinates of the object $i$, $\theta_i$ represents the heading angle of object $i$, $L_{x_{i,c}}$ and $L_{y_{i,c}}$ are the characteristic length and width of the potential field of the object $i$ with its size and speed considered [46], respectively, $E_{x_{i,c}}$ and $E_{y_{i,c}}$ are the items that incorporate perception SOTIF entropy, and $U_x$ and $U_y$ are the safety margins set to prevent hazards caused by unexpected behavior of the high uncertainty object.

On the other hand, AVs should also adhere to the road markings and not cross the road's boundary. As a result, until the lane change or overtaking instruction is sent, the AV will continue to drive in the current lane. Using a quadratic function, the potential field of road borders is designed to prevent unanticipated road crossings. The road potential field function is as follows:

$$PF_{R(X,Y)} = \begin{cases} a_q(S_{Rq}(X,Y) - D_a)^2 & S_{Rq}(X,Y) \le D_a \\ 0 & S_{Rq}(X,Y) > D_a \end{cases} \quad (12)$$

where $a_q$ represents the parameter of road boundaries $PF_R$, and $S_{Rq}$ and $D_a$ indicate the distance and safety threshold between the ego vehicle and road boundaries, respectively. The $PF_R$ will influence the ego vehicle to maintain its position in the center of the lane.

Therefore, the overall potential field can be represented as follows:

$$PF = PF_U + PF_R \quad (13)$$

Fig. 4 demonstrates an example of the overall potential field that incorporates both $PF_U$ and $PF_R$. The perception uncertainty is highlighted, i.e., the tangerine part in the figure. In summary, based on the perceptual entropy and potential field, the perceptual uncertainty is quantified, which can be used in the decision-making process.

*b. Decision-Making Paradigm Based on MPC*

First, a 3-Degree-of-Freedom (DOF) prediction model with multi-constraints is constructed to characterize the motion dynamics of the autonomous vehicle, as shown in [49], [50].

$$m(\dot{u} - vr) = F_{xT} \quad (14)$$

$$m(\dot{v} + ur) = -C_{\alpha f}(\delta_f - \frac{v + l_f r}{u}) - C_{\alpha r}(-\frac{v - l_r r}{u}) \quad (15)$$

$$I_z \dot{r} = F_{yf} l_f - F_{yr} l_r \quad (16)$$

$$\dot{X} = u\cos\phi - v\sin\phi \quad (17)$$

$$\dot{Y} = u\sin\phi + v\cos\phi \quad (18)$$

where $m$ refers to the vehicle mass, $I_z$ refers to the vehicle's moment of inertia, $l_f$ and $l_r$ represent the distance from the vehicle CG (center of gravity) to the front and rear axles, respectively, and $u$ and $v$ denote the vehicle's longitudinal and lateral velocities, respectively. The vehicle yaw rate at CG is given by $r$, $\phi$ denotes the vehicle heading angle, $X$ and $Y$ are the vehicle longitudinal and lateral positions with respect to the global coordinate, respectively, $C_{\alpha f}$ and $C_{\alpha r}$ are the cornering stiffness of the front and rear tires, respectively, $F_{xT}$ is the total longitudinal force of the tires, and $\delta_f$ is the front steering angle.

Subsequently, the model demonstrated in (15) can be denoted as the state space $f$ form with linearization and discretization techniques, i.e.:

$$\begin{aligned} x(k+1) &= A_d x(k) + B_d u(k) \\ y(k) &= C_d x(k) \end{aligned} \quad (19)$$

where $x = [u, v, r, \phi, X, Y]^T$, $u = [F_{xT}, \delta_f]^T$, $y = [Y, u]^T$, $A_d$, $B_d$ and $C_d$ denote the discrete matrices corresponding to (15), and $k$ represents the time. The details of linearization and discretization techniques and calculation process of $A_d$, $B_d$, and $C_d$ can be found in [46].

Then, based on (15) and (19), and the constructed perceptual uncertainty-aware potential field, the decision-making process can be converted into solving the following optimal control issue:

$$\begin{aligned} \min_u \sum_{k=1}^{N_p} (\|\Delta y_k\|_Q^2 + PF_k) + \sum_{k=1}^{N_c} \|u_k\|_R^2 + \|\Delta u_k\|_S^2 \\ s.\,t.\ (k = 1, ..., N_p) \\ x(t+k) = A_d x(t+k-1) + B_d u(t+k-1) \\ y(t+k) = C_d x(t+k) \\ u_{min}(t+k-1) \leq u(t+k-1) \leq u_{max}(t+k-1) \\ \Delta u_{min}(t+k-1) \leq \Delta u(t+k-1) \leq \Delta u_{max}(t+k-1) \\ u(t+k) = u(t+k-1), \forall k \geq N_c \end{aligned} \quad (20)$$

where $\Delta y_k$ represents the relative lateral position and longitudinal velocity between ego vehicle and the object, $N_p$ and $N_c$ are the prediction and control horizons, respectively, $x(t+k)$ represents the predicted state values of the system, $y(t+k)$ denotes the predicted outputs of the system over the prediction horizon, $u_{min}$ and $u_{max}$ are the lower and upper bounds of the actuator, respectively, and $\Delta u_{min}$ and $\Delta u_{max}$ are the various ranges of control variables at each time, which can be found in [46]. In the cost function shown in (20), matrices $Q$, $R$, and $S$ are the weight matrices.

## V. EXPERIMENTS AND RESULTS

For the sake of generality, this paper comes up with the study from one specific long-tail scenario related to the perceptual SOTIF problem. In this scenario, an AV travels on a straight road, while a sanitation worker sweeps on the right side of the lane with his back to the AV. Due to the similarity between the stripes and colors of sanitation workers' uniforms and traffic cones, the perception algorithm may be misled by certain trigger conditions. Once the AV erroneously identifies a sanitation worker as a traffic cone, which is considered a stationary obstacle, a collision is possible because sanitation workers may move laterally while working on the road. Therefore, it is necessary to monitor the uncertainty of perception algorithms in this scenario and account for it in algorithms that make safer decisions.

### A. Uncertainty Monitoring and Risk Quantification

As the perception algorithm of the AV, a monocular object detection algorithm is utilized in this paper. To study this sanitation-worker-traffic-cone scenario, the object detection network must be capable of recognizing at least persons and traffic cones. Consequently, the YOLOv5s object detection network is trained with the BDD dataset (70k images) and a traffic cone dataset (0.27k images) to distinguish 11 categories of objects: car, bus, truck, train, bike, motor, person, rider, traffic sign, traffic light, and traffic cone [51], [52], [53]. As the BDD dataset is much significantly larger than the traffic cone dataset, the network is trained for 60 epochs with the BDD dataset after 300 epochs of pre-training with the traffic cone dataset. On the BDD validation set, the mAP of the YOLOv5s network is 0.514, which is comparable to the performance of ResNet50 (0.499) [10]. The method of Deep Ensembles is adopted with the parameters listed in Table II.

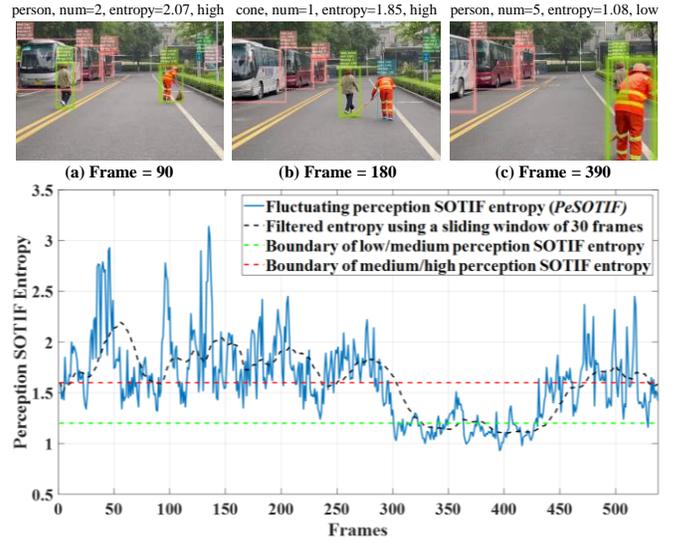

**Fig. 5.** Fluctuating perception SOTIF entropy in the test video.

The sanitation-worker-traffic-cone scenario is established in the real world, and seven videos are filmed for testing. In a local video with 538 captured frames, the PyTorch-based algorithm infers at 20 fps, and the warning accuracy reaches 97.71%. Fig. 3 depicts the framework for employing the Deep Ensembles method to estimate the epistemic uncertainty of the YOLOv5s network and the detection results of one frame in the video, while Fig. 5 depicts the fluctuating prediction entropy of the sanitation worker in the video. When the sanitation worker faces the vehicle upright or sideways, the perception algorithm can correctly identify the person with low uncertainty. However, when the sanitation worker bends or squats with the back towards the vehicle, there is a high likelihood that they will be mistaken for a traffic cone with high uncertainty. Especially when the sanitation worker is seated on the ground, even human drivers have difficulty distinguishing him.

After verifying continuous frames under the sanitation-worker-traffic-cone scenario, the effectiveness and scalability of uncertainty monitoring and risk quantification results are further verified through discrete keyframes of diverse scenarios. PeSOTIF is a labeled test dataset specifically established for vision-based probabilistic object detectors [54]. It has collected approximately 4000 objects in over 1000 keyframes of perceptual SOTIF scenarios from a variety of sources, as shown in Fig. 6. The experimental results on the PeSOTIF dataset in accordance with its evaluation protocol demonstrate that the accuracy of identifying perceptual risk reaches 90.0%, while the false alarm rate for monitoring high uncertainty objects is 10.8%. Therefore, the perception module can effectively monitor itself in most cases without being too conservative.

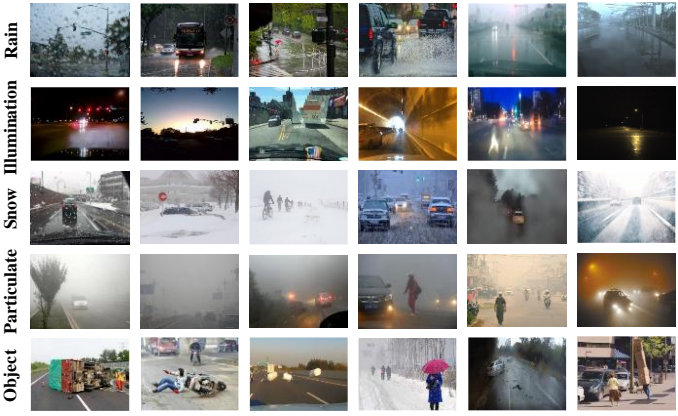

**Fig. 6.** Sample images in the PeSOTIF dataset.

In addition, the modified perception module is deployed to the AV system of the Apollo D-KIT, which utilizes an NVIDIA Geforce RTX2070S GPU, once it has been verified with a variety of perceptual SOTIF scenarios. Firstly, the Python implementation of the YOLOv5 algorithm based on PyTorch is converted to the C++ implementation based on TensorRT [55]. Secondly, the relevant necessary algorithms such as Deep Ensembles and BSASexcl are rewritten into the C++ version. Finally, the multithreaded calling method is developed to parallelize the ensemble inference of the five networks. Therefore, the algorithm that detects local videos can operate at 100 fps. In addition, the total speed can be stabilized at approximately 30 fps when acquiring and processing real-time images from the camera through CyberRT.

*B. The Performance of Risk Mitigation Through PUADM*

In the sanitation-worker-traffic-cone scenario, when the uniformed tester is normally sweeping along the road, the algorithm can correctly recognize the object as a person most of the time. However, the occasional error occurs when the tester bends over. When the tester squats or sits on the road, the algorithm frequently misidentifies him as a traffic cone, despite the fact that this situation rarely occurs in reality. As depicted in Fig. 5, the relevant data of the tester is extracted from many objects in the videos, and object tracking results are generated using a simple filtering method before being transmitted to the MPC-based decision-making algorithm downstream.

MATLAB/Simulink is utilized to implement the PUADM method introduced in Section 4.3. It establishes artificial potential fields for the perceived objects and then solves the optimal trajectory under the parameters in Table II. This paper compared two models to demonstrate the significance of considering perceptual uncertainty in subsequent tasks. The baseline model, *i.e.*, MPC-YOLO, adheres to the logic of most common modular systems, which only receive the output categories from the perceptual module, and establishes different potential field sizes for each category. The potential field of pedestrians, for instance, is the largest, followed by that of motorbikes and vehicles, and finally that of traffic cones and traffic signs, which is the smallest. PUADM receives not only the categories but also the prediction entropy as a measure of epistemic uncertainty. The artificial potential field is optimized based on uncertainty and according to the following rules: (1) low uncertainty: same as MPC-YOLO, the potential field is established based on the category; (2) Medium uncertainty: regardless of category, the potential field shall be established in accordance with the pedestrian standard; (3) High uncertainty: further expand the maximum potential field in MPC-YOLO because it is difficult to predict the behavior of unknown objects.

TABLE II
THE PREDEFINED PARAMETERS IN THE DEMONSTRATION

| Parameter | Meaning | Value |
|---|---|---|
| $T$ | The total number of sampling networks in the ensemble. | 5 |
| $C$ | The total number of categories. | 11 |
| $\theta_{aff}$ | The spatial affinity threshold in the BSASexcl algorithm. | 0.95 |
| $f_p$ | The additional penalty factor. | 0.1 |
| $\theta_{lm}$ | The threshold of entropy between low and medium levels. | 1.2 |
| $\theta_{mh}$ | The threshold of entropy between medium and high levels. | 1.6 |
| $L_{x_i,person}$ | The characteristic length of the PF of a person. | 15 m |
| $L_{y_i,person}$ | The characteristic width of the PF of a person. | 1.5 m |
| $L_{x_i,traffic\ cone}$ | The characteristic length of the PF of a traffic cone. | 8 m |
| $L_{y_i,traffic\ cone}$ | The characteristic width of the PF of a traffic cone. | 0.5 m |
| $U_x$ | The length of the safety margin. | 8 m |
| $U_y$ | The width of the safety margin. | 1 m |
| $C_{\alpha f}$ | The cornering stiffness of the front tire. | 90000 N/rad |
| $C_{\alpha r}$ | The cornering stiffness of the rear tire. | 90000 N/rad |
| $l_f$ | The distance between the mass point and the front axle. | 1.18 m |
| $l_r$ | The distance between the mass point and the rear axle. | 1.77 m |
| $m$ | The total mass of the vehicle. | 1860 kg |
| $I_z$ | The moment mass of the vehicle for Z axle. | 3438.5 kg·m$^2$ |
| $v_e$ | The expected speed of the ego vehicle. | 15 m/s |
| $v_p$ | The forward speed of the sanitation worker. | 1 m/s |
| $w_l$ | The width of the road lane. | 3.5 m |
| $x_0$ | The original distance between the person and ego vehicle. | 30 m |
| $y_0$ | The original distance between the person and the lane. | 1 m |
| $\Delta t$ | The time step of the simulation in the experiment. | 0.033 s |

The Hardware-in-the-Loop (HIL) experiment is conducted to verify the real-time performance of the decision-making algorithm synchronously. The hardware deployment is depicted in Fig. 7. In MATLAB/Simulink on the host computer, a scenario is established in which the ego vehicle advances at a constant speed of 15 m/s, and a sanitation worker sweeps near the lane centerline in front. The perception results of the videos are divided into four cases and fed into the decision-making algorithm as scenario perception information from the concurrent real-time machine. The trajectory is then generated by the MPC-based PUADM method. Afterward, the motion command is transmitted via the I/O interface to the vehicle dynamic model on the real-time machine. The real-time machine then moves in circles after transmitting the ego vehicle states of the subsequent time step to the host computer. Meanwhile, the electrical cabinet and the controller could transmit data to each other by CAN. The variables in the MATLAB/Simulink model such as ego vehicle states can be observed through the software TCS on the host computer.

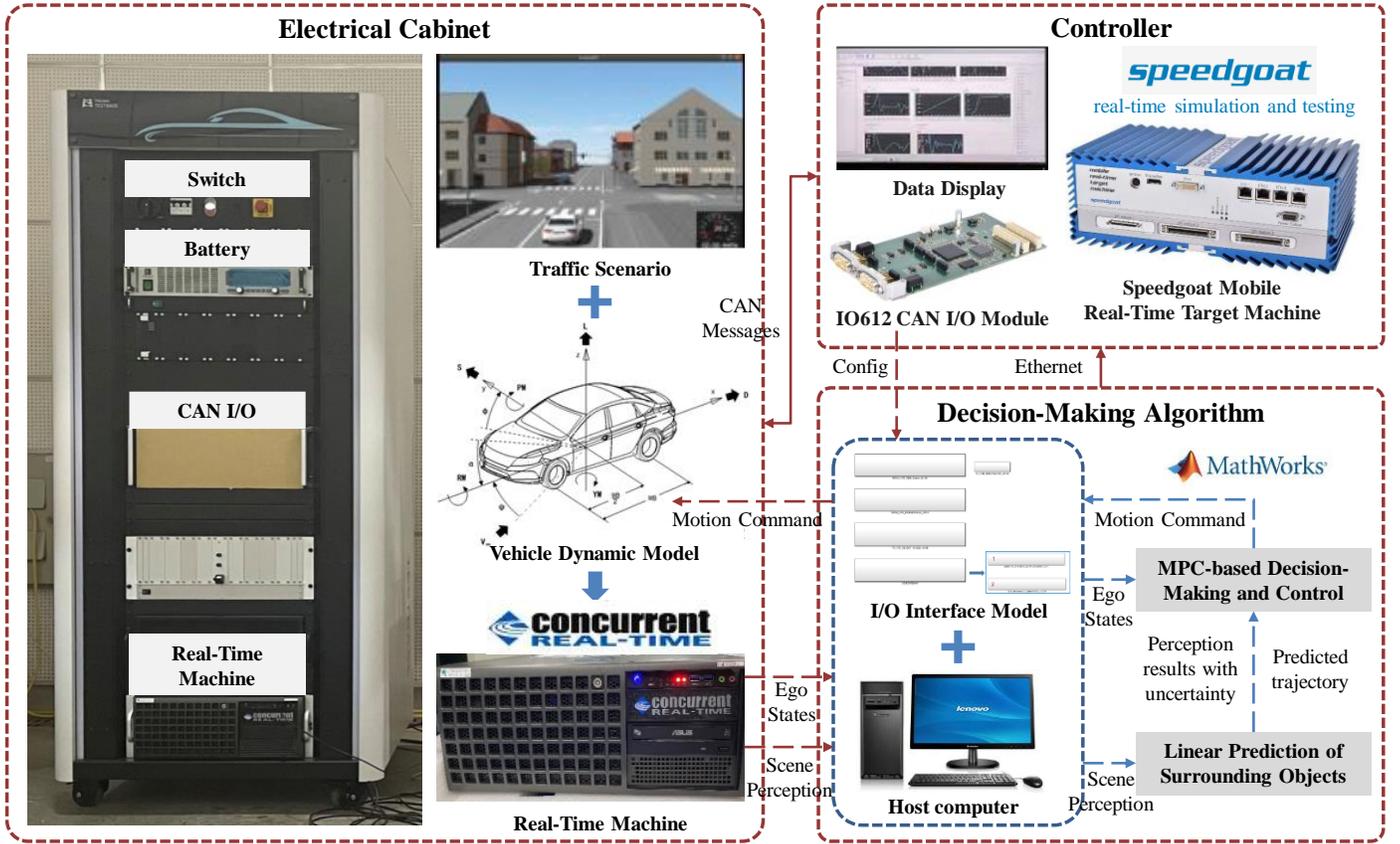

**Fig. 7.** Composition and architecture of the Hardware-in the-Loop platform.

In the HIL experiments, the vehicle dynamic MATLAB/Simulink model initially causes the ego vehicle to travel straight and at a constant speed along the road. After the ego vehicle stabilizes, another MATLAB/Simulink model constructing the scenario is initiated so that the ego vehicle can conduct trajectory planning based on perception results and uncertainty. The fixed time step is set to 0.033s, and Fig. 8 demonstrates that the real-time performance satisfies the specifications.

In case 1, the target's classification changed from person to traffic cone, while the entropy remained high throughout the process. For the baseline model, a proper evasive maneuver was made in the early stage. However, as soon as false detection occurred, the ego vehicle returned to the centerline, which is potentially dangerous. In case 2, the target's classification changed from traffic cone to person, and the filtered entropy remained high throughout the entire process. Due to the false detection in the baseline model, the ego vehicle did not evade at all and drove directly past the sanitation worker near the centerline of the road lane, which was extremely dangerous. However, the ego vehicle utilizing the PUADM model finished the complete evasive maneuver due to the high uncertainty of the target in the two cases.

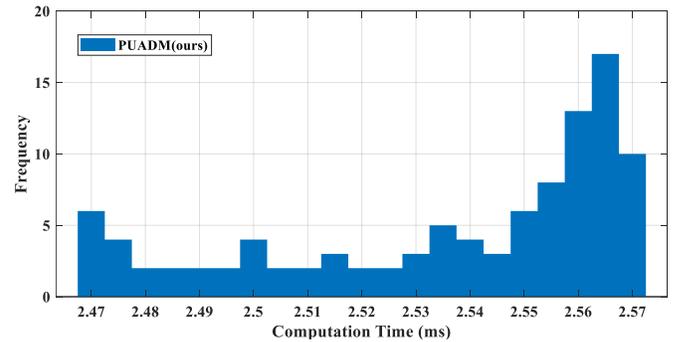

**Fig. 8.** Real-time performance of the proposed PUADM method in case 2.

In cases 3 and 4, the target was classified as a person with high and low entropy, respectively. The ego vehicle utilizing the baseline model completed a complete evasive maneuver after detecting a sanitation worker near the road's centerline. In case 3 of the PUADM model, where the target was determined to be an object with high entropy and whose behavior was difficult to predict, the ego vehicle executed a larger and earlier evasive maneuver. In case 4, where classification results were normal, the PUADM model performed similarly to the baseline model.

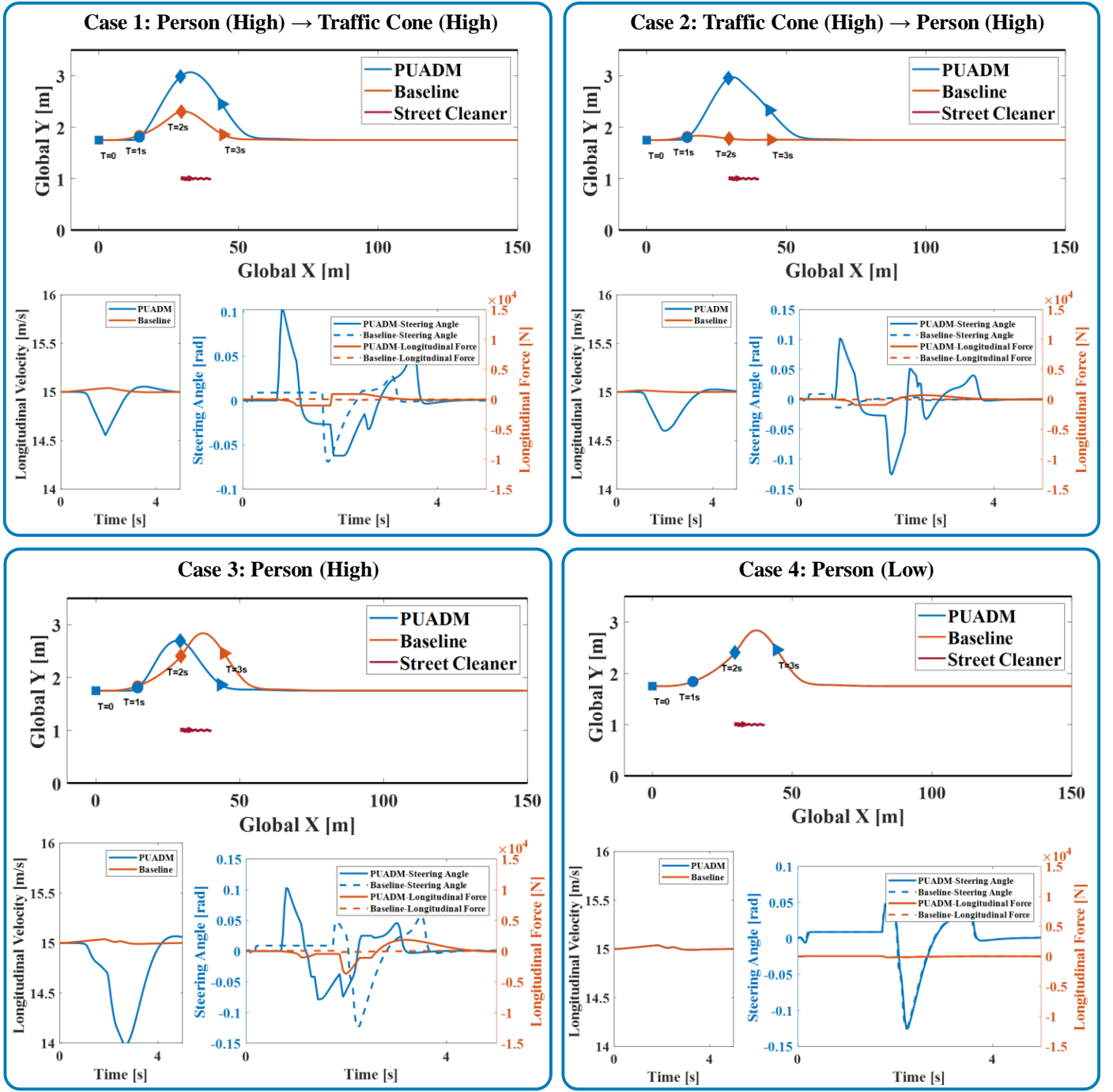

**Fig. 9.** Planning results in the test cases, including trajectory, longitudinal velocity, longitudinal force, and steering angle.

The simulation results depicted in Fig. 9 indicate that the PUADM method is safer and more aligned with the expectations of human drivers. Generally, when the uncertainty is low, the vehicle's behavior is consistent with the standard decision-making method MPC-YOLO, whereas when uncertainty is high, the vehicle will take safer actions. Specifically, the ego vehicle employing MPC-YOLO could not evade at all in case 2 due to the initial false detection. It drove directly past the pedestrian near the lane's centerline, which is extremely dangerous. Despite false detection, the ego vehicle utilizing PUADM completed the entire evasive maneuver due to the high level of uncertainty. Whether or not this method will result in overly cautious trajectory planning is dependent on whether or not the perception algorithm frequently produces high-level uncertain results. In addition, Moreover, it is related to the configuration of uncertainty thresholds and the performance of the perceptual module.

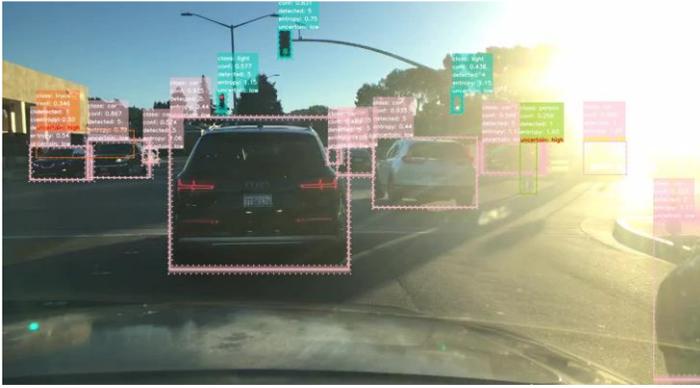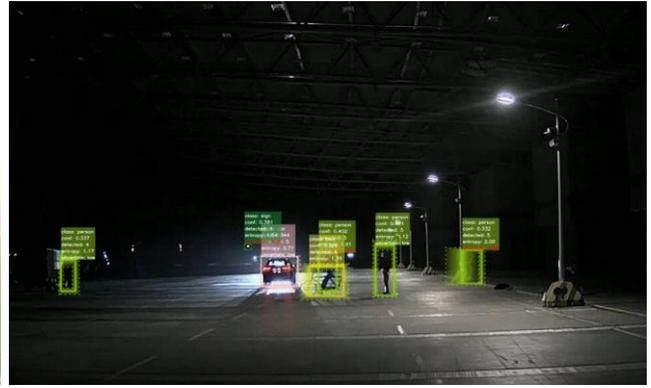

**(a) PeSOTIF-Environment-Illumination-Natural/Handcraft**

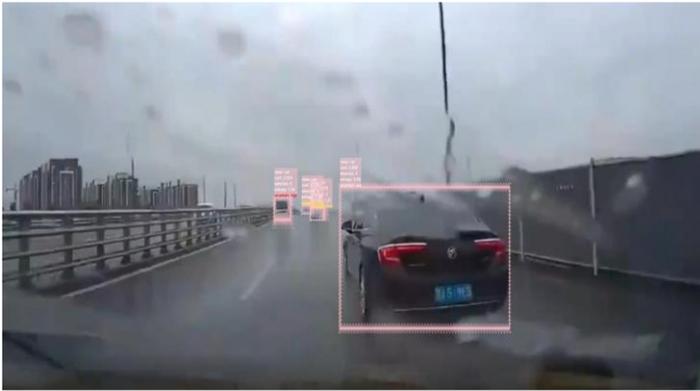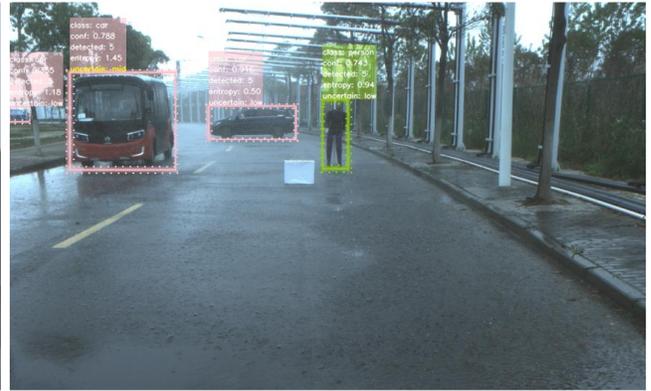

**(b) PeSOTIF-Environment-Rain-Natural/Handcraft**

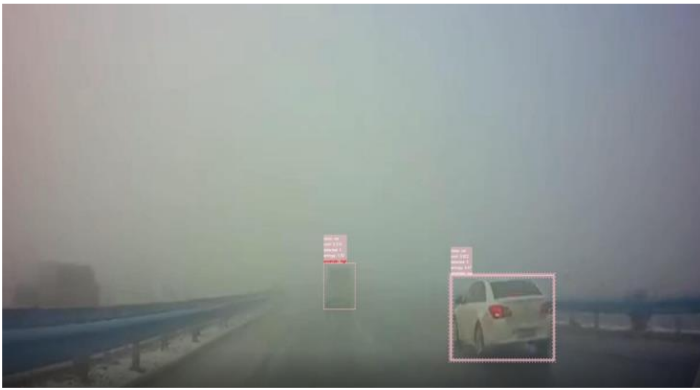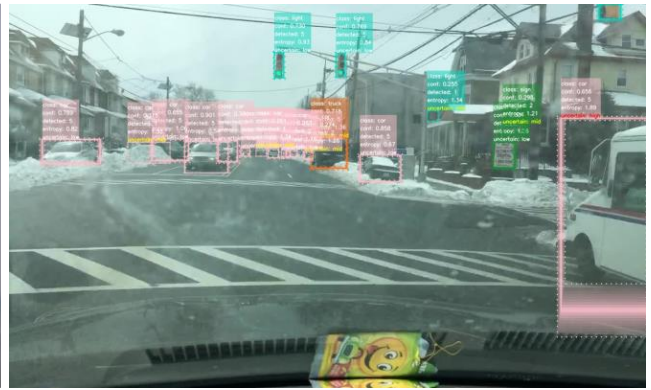

**(c) PeSOTIF-Environment-Particulate/Snow-Natural**

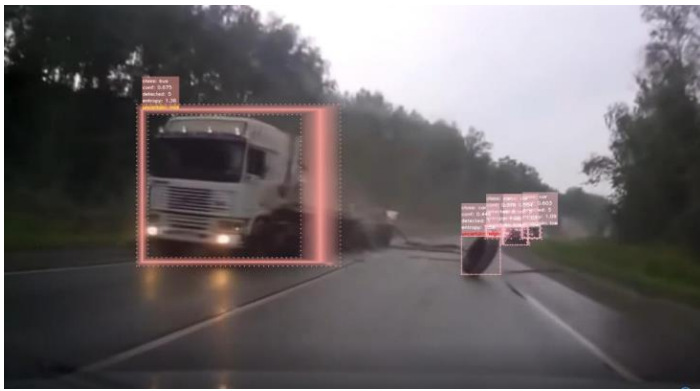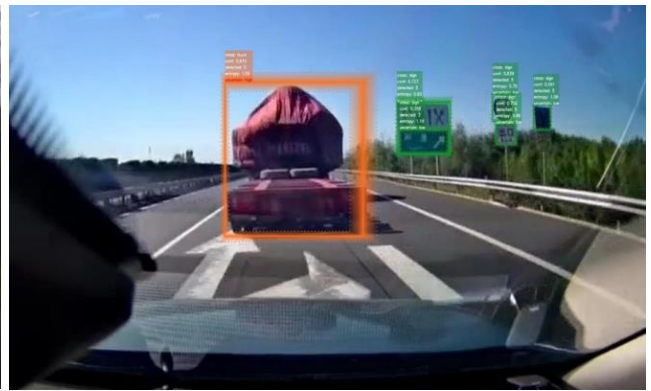

**(d) PeSOTIF-Object-Uncommon/Common-Appearance**

Fig. 10. Perception results with entropy in key frames of some typical SOTIF scenarios in the PeSOTIF dataset. Objects with low, medium, and high entropy are labeled white, yellow, and red on the last line of the box information, respectively.

## VI. Conclusion

This paper presented a systematic method for describing the overall SOTIF risk of AVs. A Self-Surveillance and Self-Adaption System was proposed to monitor, quantify, and mitigate the SOTIF risk. Then, a demonstration system was developed by estimating the uncertainty of YOLOv5 to monitor the perceptual risk, quantifying the uncertainty as SOTIF entropy, and mitigating the entropy via a Perceptual Uncertainty-Aware Decision-Making (PUADM) technique. Diverse perceptual SOTIF scenarios were evaluated to confirm the performance of the perception module. Afterward, Hardware-in-the-Loop (HIL) experiments were conducted to confirm the system's real-time performance and effectiveness. Experimental experiments demonstrate that the estimated perception SOTIF entropy is dependable and the PUADM method is safer than the baseline method without being too conservative.

This paper quantified only the perception SOTIF entropy, disregarding the prediction SOTIF entropy and the planning SOTIF entropy, as well as the potential impact of prediction errors on planning in the demonstration. The system will become more complex if learning-based prediction and planning algorithms such as Long Short-Term Memory and Reinforcement Learning are considered. Future research on the overall SOTIF risk will focus on how to quantify the SOTIF entropy from learning-based prediction and planning algorithms and how to sort out the coupling relationship between them. Although the performance of the perception module and the PUADM method has been verified through simulation experiments on the Apollo platform and the HIL platform, respectively, this system has not been exposed to the actual perceptual SOTIF scenarios. The Self-Surveillance and Self-Adaption System will be deployed to an autonomous truck to conduct field experiments to verify its performance soon after.

## Appendix

Fig. 10 shows some perception results of the probabilistic object detector based on YOLOv5 and Deep Ensembles on the PeSOTIF dataset. Although the truck with a strange appearance in (d) is correctly classified, it is meaningful to output high entropy.

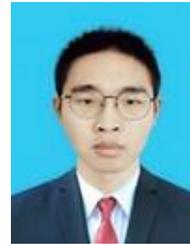

**Liang Peng** received the B.Eng. degree in Vehicle Engineering from the School of Vehicle and Mobility of Tsinghua University, Beijing, China, in 2020. He is currently working toward a Ph.D. degree in Vehicle Engineering at Tsinghua University. He is a member of the Tsinghua Intelligent Vehicle Design And Safety (IVDAS) Research Institute and is supervised by Prof. Jun Li and Hong Wang. His research interests include the evaluation and uncertainty analysis of perceptual algorithms and the safety of autonomous driving vehicles.

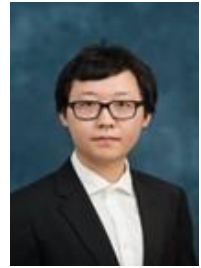

**Boqi Li** received the B.S. degree in mechanical engineering from the University of Illinois Urbana–Champaign, Champaign, IL, USA, in 2015, and the M.S. degree in mechanical engineering from Stanford University, Stanford, CA, USA, in 2017. He is currently pursuing the Ph.D. degree in mechanical engineering with the University of Michigan, Ann Arbor, MI, USA.

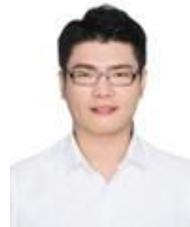

**Wenhao Yu** is currently a research associate at school of Vehicle and mobility with Tsinghua University. He received his Ph.D. degree in Jiangsu University in China at 2020. His research focuses on the decision-making, path planning and following control of autonomous vehicles,Model Predictive Control and Safety of The Intended Functionality of autonomous vehicles.

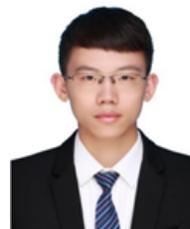

**Kai Yang** is currently pursuing a Ph.D. degree in the Department of Automotive Engineering, Chongqing University, Chongqing, China. He received a B.E. degree in vehicle engineering from the Wuhan University of Technology in 2018. He researches as a Joint Ph.D. Student in School of Vehicle and Mobility, Tsinghua University, Beijing, China. His research interests focus on motion prediction and decision-making of autonomous vehicles.

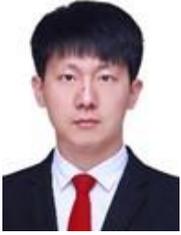
**Wenbo Shao** received his B.E. degree in vehicle engineering from Tsinghua University, Beijing, China, in 2019. He is currently working toward a Ph.D. degree in Mechanical Engineering at Tsinghua University. He is a member of the Tsinghua Intelligent Vehicle Design And Safety Research Institute (IVDAS) and is supervised by Professor Jun Li and Associate Research Professor Hong Wang. His research interests include safety of the intended functionality of autonomous driving, prediction, decision-making, uncertainty theory, and applications.

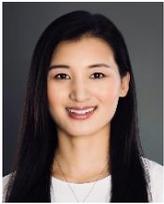
**Hong Wang** (Member, IEEE) is currently a Research Associate of Mechanical and Mechatronics Engineering with the University of Waterloo. She received her Ph.D. degree in Beijing Institute of Technology in China in 2015. Her research focuses on the path planning control and ethical decision making for autonomous vehicles and component sizing, modeling of hybrid powertrains and power management control strategies design for Hybrid electric vehicles; intelligent control theory and application.
+86 18611819756
hong_wang@tsinghua.edu.cn